\documentclass[review]{elsarticle}
\usepackage{times}  % DO NOT CHANGE THIS
\usepackage{helvet}  % DO NOT CHANGE THIS
\usepackage{courier}  % DO NOT CHANGE THIS
\usepackage[hyphens]{url}  % DO NOT CHANGE THIS
\usepackage{graphicx} % DO NOT CHANGE THIS
\usepackage{natbib}  % DO NOT CHANGE THIS AND DO NOT ADD ANY OPTIONS TO IT
\usepackage{caption} % DO NOT CHANGE THIS AND DO NOT ADD ANY OPTIONS TO IT
\usepackage{amsfonts}

\usepackage{algorithm}
\usepackage{algorithmic}
\usepackage{amsmath}
\usepackage{xcolor}
\usepackage{enumitem}
\setlist[itemize]{align=parleft,left=0pt..1em}

\usepackage{array}
\usepackage{multirow}
\usepackage{booktabs}
\usepackage{url}

\usepackage{newfloat}
\usepackage{listings}
\usepackage{lineno,hyperref}
\modulolinenumbers[5]

\journal{Biomedical Signal Processing and Control}

\begin{document}

\begin{frontmatter}

\title{HUR-MACL: High-Uncertainty Region-Guided Multi-Architecture Collaborative Learning for Head and Neck Multi-Organ Segmentation}

%% 作者信息（使用†标记共同第一作者，*标记通讯作者）
\author[mymainaddress,mysecondaryaddress]{Xiaoyu Liu$^{\dag}$}
\author[mymainaddress,mysecondaryaddress]{Siwen Wei$^{\dag}$}
\author[mymainaddress,mysecondaryaddress]{Linhao Qu$^{\dag}$}
\author[mythirdaddress]{Mingyuan Pan}
\author[mymainaddress,mysecondaryaddress]{Chengsheng Zhang}
% 通讯作者用*标记
\author[mymainaddress,mysecondaryaddress]{Yonghong Shi$^{*}$}
\author[mymainaddress,mysecondaryaddress]{Zhijian Song$^{*}$}

% 1. 共同第一作者脚注
\fntext[1]{$^{\dag}$These authors contributed equally to this work.}

% 2. 通讯作者脚注（标准格式）
\fntext[2]{$^{*}$Corresponding authors. Email addresses: yonghong.shi@fudan.edu.cn (Yonghong Shi), zjsong@fudan.edu.cn (Zhijian Song)}

% 作者单位地址（考虑合并相关地址）
\address[mymainaddress]{Digital Medical Research Center, School of Basic Medical Science, Fudan University, Shanghai 200032, China}
\address[mysecondaryaddress]{Shanghai Key Lab of Medical Image Computing and Computer Assisted Intervention, Shanghai 200032, China}
\address[mythirdaddress]{Radiation Oncology Center, Huashan Hospital Affiliated to Fudan University, Shanghai 200032, China}

\begin{abstract}
Accurate segmentation of organs at risk in the head and neck is essential for radiation therapy, yet deep learning models often fail on small, complexly shaped organs. While hybrid architectures that combine different models show promise, they typically just concatenate features without exploiting the unique strengths of each component. This results in functional overlap and limited segmentation accuracy. To address these issues, we propose a \underline{h}igh-\underline{u}ncertainty \underline{r}egion-guided \underline{m}ulti-\underline{a}rchitecture \underline{c}ollaborative \underline{l}earning (HUR-MACL) model for multi-organ segmentation in the head and neck. This model adaptively identifies high uncertainty regions using a convolutional neural network, and for these regions, Vision Mamba as well as Deformable CNN are utilized to jointly improve their segmentation accuracy. Additionally, a heterogeneous feature distillation loss was proposed to promote collaborative learning between the two architectures in high uncertainty regions to further enhance performance. Our method achieves SOTA results on two public datasets and one private dataset.
\end{abstract}


\end{frontmatter}

% \linenumbers

\section{Introduction}

Head and neck cancers are of significant concern due to their high incidence and mortality rates \cite{1}. Radiation therapy as the primary treatment method, relies on accurately targeting tumors while minimizing radiation damage to surrounding Organs at Risk (OARs) \cite{2}. However, manually contouring tumors and OARs on CT images is time-consuming and prone to variability across patients, potentially compromising treatment efficacy \cite{3}.

\begin{figure}[t!]  
\centering
  \includegraphics[width=\textwidth]{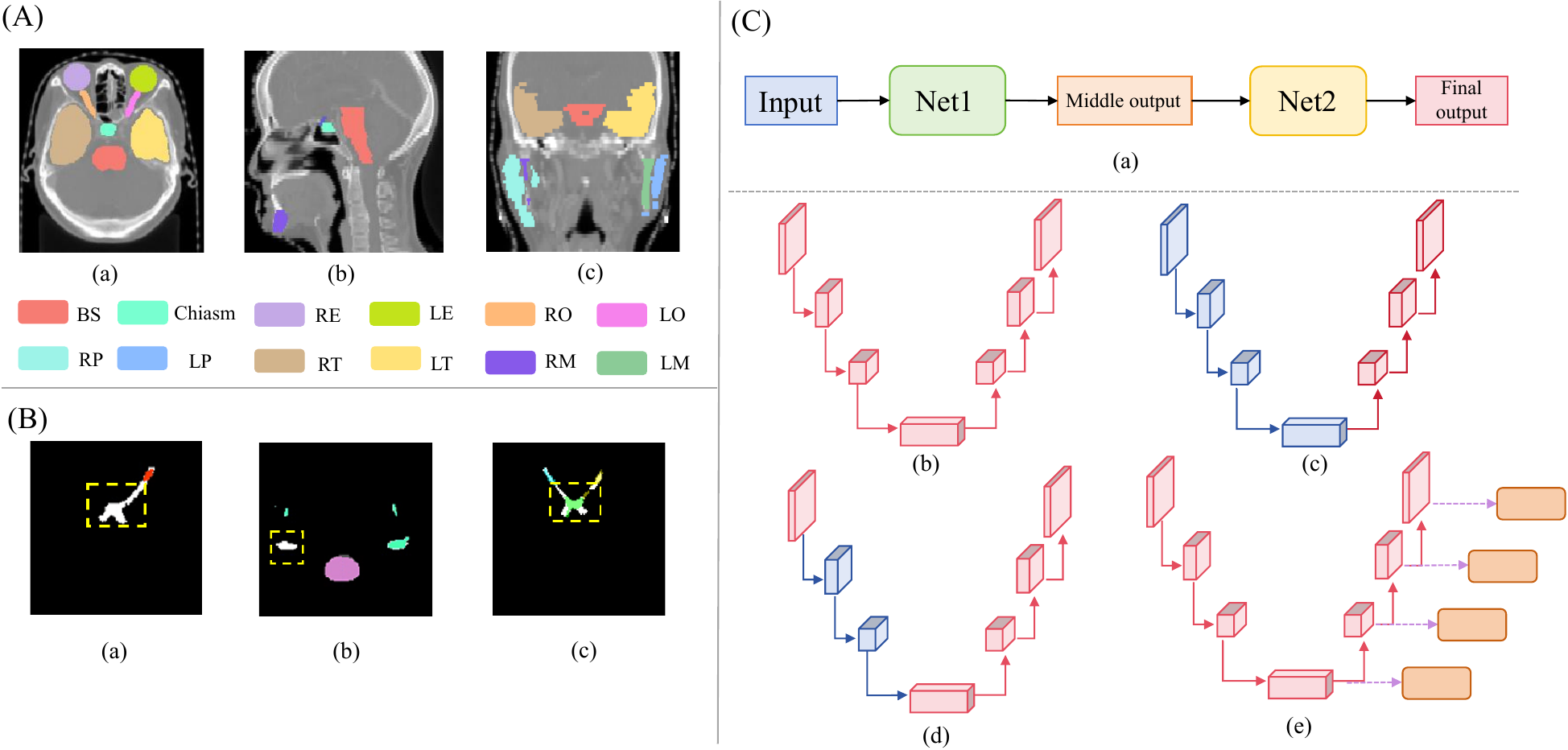} 
  \caption{(A). Schematic diagram of HNOARs. BS (Brainstem), RE (Right Eye), LE (Left Eye), RON (Right Optic Nerve), LON (Left Optic Nerve), RPG (Right Parotid Gland), LPG (Left Parotid Gland), RT (Right Temporal lobe), LT (Left Temporal lobe), RM (Right Mandible), LM (Left Mandible). (B). Segmentation error cases, where white denotes the ground truth and color denotes the predictions. (C). Segmentation architectures, where different colors represent different networks.}
  \label{fig:1}
\end{figure}

Recently, deep learning-based automated segmentation of head and neck OARs (HNOARs) has emerged as a promising solution \cite{4}. However, existing methods often struggle to meet the stringent requirements for clinical use, facing two primary anatomical challenges. First, the head and neck region is characterized by the dense spatial arrangement of numerous OARs, as shown in {Fig.~\ref{fig:1} (A)}, which requires models to understand global context and inter-organ relationship. Second, many individual organs, such as the optic chiasm highlighted in {Fig.~\ref{fig:1} (B)}, possess intrinsically complex and irregular morphologies, demanding the precise capture of fine-grained local features. Consequently, a key limitation of current methods is their inability to effectively reconcile these demands, that is, they fail to adequately capture both the global anatomical layout and the local boundary details simultaneously. This leads to suboptimal accuracy, particularly for small or intricately shaped structures \cite{5} that are crucial for treatment planning \cite{6}.

Many studies have shown that advanced network design can help address the above challenges in HNOARs segmentation \cite{7,8,9,10,11}. Existing methods are generally categorized into Multi-Stage and Single-Stage ({Fig.~\ref{fig:1} (C)}). The core idea of Multi-Stage methods is “divide and conquer” or “coarse to fine”, and their workflow is usually as follows: first, a network (first stage) performs preliminary segmentation (such as segmenting large and easily recognizable organs), and then inputs its output or features as prior information into the second network (second stage) to achieve fine segmentation of small and difficult-to-process organs \cite{6,12}. For example, MHL-Net \cite{6} first uses MsegNet to segment large organs, then combines its outputs with Anatomical Prior Probability Map (APPM) to locate and segment small organs. Similarly, FocusNet \cite{12} first segments large organs using S-Net, then locates small organs from S-Net's features using SOL-Net, and finally achieves precise segmentation of these small organs using SOS-Net. However, multi-stage methods also have inherent drawbacks such as error accumulation, high model complexity, and slow computation speed.

Unlike multi-stage methods, Single-Stage methods utilize a single network to perform end-to-end segmentation of HNOARs \cite{5}. In terms of network architecture, these models can be further classified into single architecture and hybrid architecture. The single architecture model only uses one core component, such as CNN \cite{13,14}, Transformer \cite{15}, Mamba \cite{16}. However, single architectures are limited by their inherent constraints, hindering further improvements. For example, CNN-based models, though computationally efficient, struggle to capture global features; Transformer and Mamba-based models excel at capturing global features but are less effective at local feature extraction. To address this issue, hybrid models such as TransUNet \cite{17} and Mamba-UNet \cite{18} complement the strengths of different architectures, often yielding superior performance, making them the mainstream choice in the field \cite{19}.

Despite the progress of hybrid models, most hybrid models simply concatenate features from different architectures without adaptively combining them based on the specific characteristics of organs in CT images. For larger and regularly shaped organs (e.g., brainstem, parotid glands), CNNs alone can achieve satisfactory segmentation accuracy. However, for smaller and more complex organs (e.g., spinal cord, chiasm), it is necessary to rely on the collaborative work of multiple architectures to achieve effective segmentation \cite{20,21}. The current strategy of integrating multiple architecture feature indiscriminately in all regions not only fails to fully leverage the unique advantages of each architecture, but also leads to feature redundancy and increased computational costs, ultimately limiting the improvement of segmentation accuracy \cite{22}. Besides, most existing methods were designed for datasets in other areas such as abdominal, and their performance is often unsatisfactory when directly transferred to the more complex anatomical structure of the head and neck. Therefore, there is an urgent need to develop dedicated multi-organ segmentation methods for HNOARs.

To address these challenges, we propose a novel model called High Uncertainty Region Guided Multi-Architecture Collaborative Learning (HUR-MACL), specifically designed for multi organ segmentation tasks in the head and neck region. HUR-MACL aims to adaptively identify and focus on high uncertainty areas in segmentation, and strategically integrate the advantages of various architectures, thereby improving the efficiency and accuracy of segmentation. The model consists of three main components: \textbf{High uncertainty region mining}. CNN is first used for preliminary segmentation. Subsequently, by calculating the uncertainty of the CNN output, the high uncertainty region is able to be mined. The typical features of these regions are complex organ morphology or the absence of global contextual information during segmentation. \textbf{Multi-architecture collaborative segmentation}. For the high uncertainty areas identified, we adopt a hybrid architecture for focused supervision and optimization. Specifically, we use Vision Mamba \cite{23} to capture global features over long distances, while utilizing Deformable CNN \cite{24,25,26} to adaptively handle targets with irregular shapes and sizes. \textbf{Heterogeneous feature distillation}. We introduce a novel heterogeneous feature distillation loss function to promote mutual learning between Vision Mamba and Deformable CNN architectures, thereby synergistically enhancing the segmentation performance of the model in high uncertainty regions. 

\textbf{The main contributions of this paper are as follows:}

\begin{itemize}
    \setlength{\itemindent}{2em}  % 控制缩进大小
    \item {Propose a multi-architecture collaborative learning model guided by high-uncertainty regions for accurate head and neck multi-organ segmentation.}
    \item {Design a novel heterogeneous feature distillation loss to enhance collaboration between Vision Mamba and Deformable CNN.}
    \item {Achieve SOTA performance on two public and one private datasets, outperforming eight existing methods.}
\end{itemize}

\section{Related Work}

\subsection{Medical Image Segmentation}
\label{sec21}
Medical image segmentation plays a crucial role in clinical diagnostic and treatment planning by accurately depicting anatomical structures and pathological regions. Recent studies highlight that advancements in deep learning-based network design significantly improve segmentation accuracy and efficiency \cite{7,8,9,10,11,liu2024deep}. Various approaches have been explored, with some models relying on a single architecture type, such as Convolutional Neural Networks (CNNs) \cite{13,14}, known for their local feature extraction capabilities; Transformers \cite{15}, which excel in capturing long-range dependencies; or State Space Sequence Models like Mamba \cite{16}, which offer efficient long-sequence modeling. Alternatively, hybrid architectures have gained attention by integrating the complementary advantages of different frameworks. For instance, TransUNet \cite{17} combines Transformer with U-Net’s hierarchical structure to enhance global context awareness, while Mamba-UNet \cite{18} integrates Mamba’s selective state space mechanisms with U-Net for improved computational efficiency in medical images.

\subsection{OARs Segmentation for Head and Neck Region}
\label{sec22}

{The methodology for multi-organ segmentation in the head and neck region has witnessed a paradigm shift from multi-stage strategies to end-to-end hybrid architectures.}

{Early approaches often employed multi-stage designs to address the complexity of anatomical structures in this area. For instance, FocusNet \cite{12} adopted a three-stage cascaded network to sequentially perform coarse segmentation, localization of small organs, and fine-grained optimization. Similarly, MHL-Net \cite{6} leveraged large-organ segmentation to generate prior maps guiding the segmentation of small organs. These methods improved accuracy to some extent by decomposing tasks, but their multi-stage architectures resulted in complex training, low inference efficiency, and difficulties in integrated optimization.}

{To mitigate these issues, the paradigm shifted toward single-stage end-to-end models, focusing on the hybrid architectural integration\cite{chen2021transunet,18}. However, a critical bottleneck remains at the feature fusion stage. While backbones have evolved, the mechanisms for aggregating multiscale features remain largely static and rudimentary (e.g., element-wise concatenation or summation). In the head and neck region, where organs exhibit extreme disparities in scale (e.g., optic nerves vs. parotid glands) and low inter-class contrast, these general fusion strategies fail to dynamically prioritize relevant features. They cannot adaptively weigh the trade-off between semantic context for large organs and boundary details for small structures.}

{To bridge this gap, this paper proposes a novel dynamic adaptive fusion paradigm, enabling the network to automatically adjust fusion strategies based on input features, thereby more fully leveraging the advantages of different architectures to enhance the accuracy and robustness of multi-organ segmentation in the head and neck region.}

\section{Method}
\subsection{Framework Overview}
\begin{figure}[t!]  
\centering
  \includegraphics[width=\textwidth]{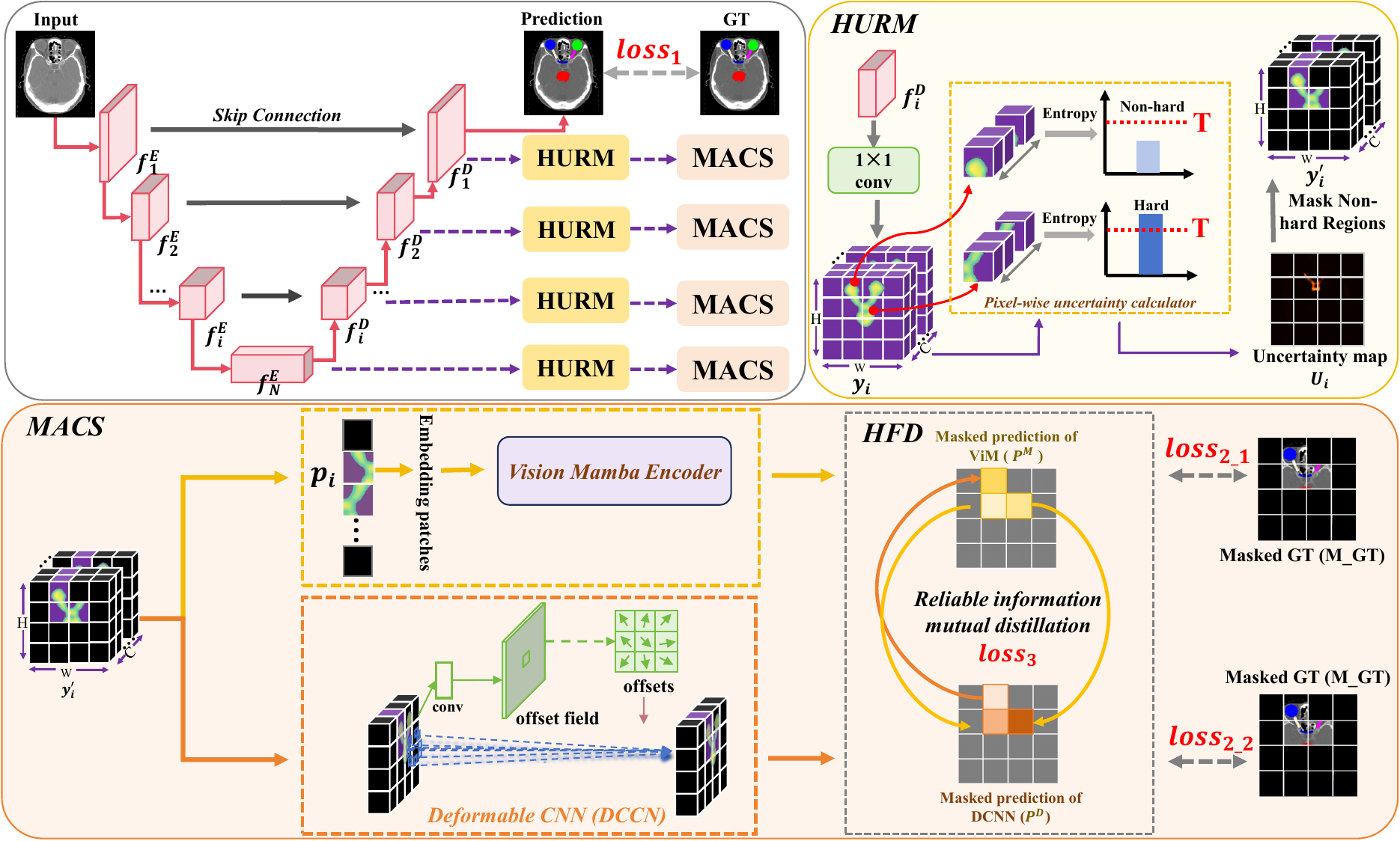} 
  \caption{The overall framework of HUR-MACL, which involves hard region mining to mask non-hard regions, followed by multi-architecture collaborative segmentation (using ViM and DCNN) on the masked feature map, and heterogeneous feature distillation module is introduced to exchange reliable information in hard regions.}
  \label{fig:2}
\end{figure}

We present a novel high uncertainty region guided multi-architecture collaborative segmentation model (HUR-MACL) tailored for HNOARs segmentation. As shown in {Fig.~\ref{fig:2}}, HUR-MACL is based on a U-Net architecture and innovatively integrates three core components: High Uncertainty Region Mining (HURM), Multi-Architecture Collaborative Segmentation (MACS), and Heterogeneous Feature Distillation (HFD). The workflow of the model is as follows: Firstly, CT images are input into the U-Net backbone network to generate predictions and calculate segmentation loss with ground truth (GT) ($loss_{1}$). In its decoding stage, the HURM module we designed will generate a pixel-by-pixel uncertainty map based on the decoder feature $f_i^D$ to adaptively identify the "hard regions" (i.e. high uncertainty regions) in the segmentation task. Subsequently, these identified hard areas will be sent to the MACS module for expert level fine processing. The MACS module consists of two parallel, structurally heterogeneous branches: one based on Vision Mamba (ViM) and the other based on Deformable Convolutional Networks (DCNN), aimed at leveraging their complementary advantages in feature extraction. In order to achieve effective collaboration between the two architectures, we further introduced the HFD mechanism, which forces ViM and DCNN to learn and supervise each other in high uncertainty regions through feature distillation. In this way, HUR-MACL could target and optimize challenging areas, thereby improving overall segmentation accuracy.

\subsection{High Uncertainty Region Mining (HURM)}

{High} Uncertainty Region Mining is performed by calculating the uncertainty of the output at each decoder layer, with high-uncertainty areas identified as hard regions. Specifically, the output feature map {$f_{i}^{D}$} from each layer of the decoder is first transformed into {$y_{i} \in \mathbb{R}^{H \times W \times C}$} through a $1 \times 1$ convolution, where $(H,W)$ represents the size of the feature map and $C$ is the channel dimension. The uncertainty map $U_i(h,w)$ is then computed for each pixel using the normalized softmax entropy in Eq.~\eqref{eq:1}, where {$y_i(h,w,c)$} denotes the predicted probability for class $c$ at position $(h,w)$.

Next, a threshold $T$ is selected to binarize $U_i(h,w)$, resulting in a binary map {$M_i \in {(0,1)}^{H \times W}$}. By multiplying $M_i$ with {$y_i$}, pixels with entropy higher than $T$ are retained, thus masking the non-hard regions and obtaining the hard regions, which finally produces the masked feature map {$y_{i}^{\prime}$}.

\begin{equation}
\scriptsize
U_i(h, w) = - \frac{\sum_{c \in C} \left( y_i(h, w, c) \cdot \log \left( y_i(h, w, c) \right) \right)}{\log |C|}
\label{eq:1}
\end{equation}

\subsection{Multi-Architecture Collaborative Segmentation (MACS)}
\begin{figure}[t!]  
\centering
 \includegraphics[width=\textwidth]{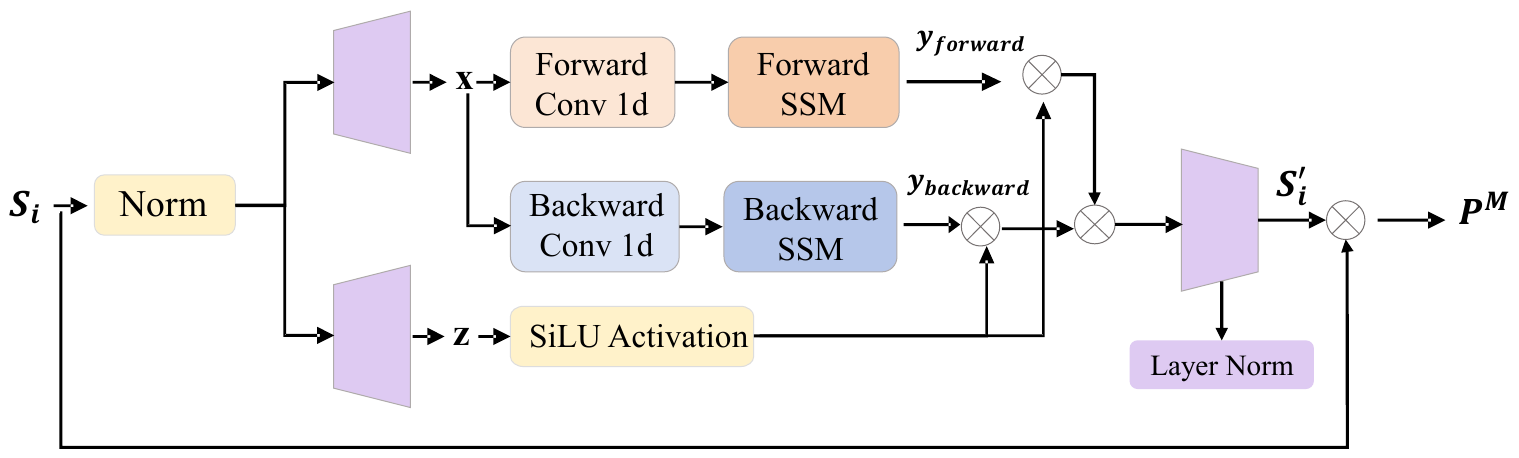} 
  \caption{The overall framework of Vision Mamba Encoder.}
  \label{fig:3}
\end{figure}

The multi-architecture collaborative segmentation involves using the ViM encoder and DCNN to segment $y_i^{\prime}$ separately, as well as mutual distillation between them in hard regions. First, $y_i^{\prime}$ is fed into the ViM encoder to extract features, capitalizing on its superior capability for global feature extraction. Specifically, $y_i^{\prime}$ is transformed into a flattened 2D image patch $p_i \in \mathbb{R}^{L \times D}$, where $D = (P_h \times P_w \times C)$ represents the size of the image patches, and $L$ represents the number of image patches. After linear mapping to obtain embedding vectors and adding position embeddings, the initial sequence $S_i$ is constructed. As shown in {Fig.~\ref{fig:3}}, $S_i$ is normalized through a normalization layer and then linearly projected into two vectors $x$ and $z$ of dimension $E$. The vector $x$ undergoes forward and backward processing, computing $y_{forward}$ and $y_{backward}$ through State Space Model (SSM). Subsequently, the feature vector $z$ serves as a gating signal to modulate the information flow from both processing paths through element-wise multiplication. Specifically, $z$ is used to gate $y_{\text{forward}}$ and $y_{\text{backward}}$, which are then combined to produce the output token sequence $S_i^{\prime}$. The complete computation can be expressed as:

{\begin{equation}
S_i^{\prime} = \text{LayerNorm}\big( \text{SiLU}(z) \odot y_{forward} + \text{SiLU}(z) \odot y_{backward} \big)
\end{equation}}

{where $\text{SiLU}(\cdot)$ denotes the Sigmoid-weighted Linear Unit activation function, $\odot$ represents element-wise multiplication, and $\text{LayerNorm}(\cdot)$ is the layer normalization operation. $S_i^{\prime}$ is then mapped back to the original image size to obtain the masked prediction of ViM ($P^M$), which is used to compute the segmentation loss ($loss_{2\_1}$) with the masked ground truth ($M\_GT$).}

Next, $y_i^{\prime}$ is fed into the DCNN encoder to extract features, leveraging its ability to capture specialized shape features. Unlike basic convolution, which samples features at fixed grid positions, DCNN learns an additional set of offsets before the convolution operation. These offsets adjust the positions of the sampling points, enabling the kernel to adaptively align with the shape variations of the target. The masked prediction $P^D$ is obtained after passing through the DCNN, and the segmentation loss ($loss_{2\_2}$) is computed between $P^D$ and $M\_GT$.

Thus, the additional segmentation loss introduced by ViM and DCNN for hard regions can be summarized by the following formula:
\begin{equation}
loss_{2} = loss_{2\_1} + loss_{2\_2}
\label{eq:2}
\end{equation}

\subsection{Heterogeneous Feature Distillation (HFD)}
To improve the feature extraction capabilities of ViM and DCNN in hard regions and promote mutual learning, we introduce a Heterogeneous Feature Distillation (HFD) module. Since ViM and DCNN exhibit distinct segmentation strengths across different areas of the hard regions, inspired by \cite{27}, we dynamically select reliable information for mutual knowledge distillation (KD) between these two models, thereby leveraging their complementary strengths. As shown in {Fig.~\ref{fig:2}}, the gray squares in ${P}^{M}$ and ${P}^{D}$ represent the masked regions. In the unmasked regions, darker colors indicate higher cross-entropy loss, implying less reliable segmentation. Therefore, we transfer reliable knowledge from the lighter-colored regions to the darker-colored regions.

Specifically, we use a matrix {$m(h,w) \in {(0,1)}^{H \times W}$} to determine the KD direction for each pixel. When the cross-entropy loss of a pixel in the ViM-based model is lower than that in the DCNN-based model, $m(h,w)$ is set to $1$, enabling knowledge transfer from ViM to DCNN. Conversely, when $m(h,w)$ is $0$, the transfer direction is reversed. After determining the KD direction for each pixel, the two models can exchange reliable pixel-level information. The specific loss functions for the two models are as follows:

\begin{equation}
\scriptsize
L_P^M = \frac{1}{H \times W - M - G} \sum_{i=1}^{H} \sum_{j=1}^{W} \left( (1 - m(i, j)) \cdot {KL}(P^M(i, j) \parallel P^D(i, j)) \right)
\label{eq:3}
\end{equation}

\begin{equation}
\scriptsize
L_P^D = \frac{1}{M} \sum_{i=1}^{H} \sum_{j=1}^{W} \left( m(i, j) \cdot \mathrm{KL}(P^M(i, j) \parallel P^D(i, j)) \right)
\label{eq:4}
\end{equation}
Here, $M = \sum_{i=1}^{H} \sum_{j=1}^{W} m(i, j)$ denotes the total number of pixels where the value of $m(i, j)$ is 1, representing the set of regions identified as hard regions. The set $G$ represents the collection of regions that are masked.

The reliable region mutual distillation loss between ViM and DCNN can be formulated as follows:
\begin{equation}
loss_{3} = L_{P}^{M} + L_{P}^{D}
\label{eq:5}
\end{equation}

\subsection{Overall Loss}
The overall loss function is composed of three components. The primary segmentation loss is calculated using the U-Net segmentation results and GT ($loss_{1}$). An additional segmentation loss generated by ViM and DCNN for hard regions ($loss_{2}$). The heterogeneous feature distillation loss fosters mutual learning between ViM and DCNN ($loss_{3}$). These components are weighted and summed to optimize overall segmentation performance. As shown in the following formula, the hyperparameters $\alpha$ and $\beta$ are introduced to balance the weights of $loss_{2}$ and $loss_{3}$, respectively.:
\begin{equation}
Loss = loss_{1} + \alpha \cdot loss_{2} + \beta \cdot loss_{3}
\label{eq:6}
\end{equation}

\section{Experiment and Results}
\subsection{Datasets}
To comprehensively evaluate our proposed method, we conducted experiments on three diverse datasets: two publicly available datasets, PDDCA (48 cases) and StructSeg (50 cases), as well as a private dataset containing 118 clinical cases (Inhouse dataset). The PDDCA and StructSeg datasets focus on segmenting head and neck organs-at-risk (OARs), including the brainstem (BS), mandible, chiasm, left and right optic nerves (LON, RON), left and right parotid glands (LPG, RPG), and Right Lens(RL). The Inhouse dataset further expands the range of anatomical structures evaluated by including annotations for the spinal cord in addition to the aforementioned OARs, making our evaluation more realistic in clinical radiotherapy scenarios.

\subsection{Comparison Methods}
To assess the effectiveness of our method, we compared it with a range of baseline and state-of-the-art segmentation approaches, including: Single-architecture methods: U-Net and nnUNet, representing widely used convolutional frameworks; Hybrid-architecture methods: TransUNet, Missformer, Mamba-UNet, and Swin-Mamba, which incorporate transformer or Mamba-based modules to enhance long-range feature modeling; Multi-stage method: MHL-Net and FocusNet, which refine segmentation outputs through cascaded stages. All comparison models were re-implemented or trained using their official repositories with identical preprocessing and data splits for fair evaluation.

\subsection{Experimental Setup}
\label{sec42}
All experiments were conducted using PyTorch 1.10.2 on a workstation equipped with an NVIDIA GeForce RTX 4090 GPU. The models were trained using stochastic gradient descent (SGD) with Nesterov momentum (µ = 0.99), an initial learning rate of 0.001, and a learning rate decay schedule applied throughout training. Model performance was quantitatively evaluated using two commonly adopted metrics in medical image segmentation: Dice Similarity Coefficient (DSC) and Average Symmetric Surface Distance (ASSD).

\subsection{Main Results}
\label{sec43}
{Tables~\ref{tab:1}-\ref{tab:3}} present comprehensive quantitative evaluations across three datasets, demonstrating the consistent superiority of our method over all comparative approaches. The analysis reveals several key observations: (1) CNN-based methods exhibit strong performance for regularly-shaped, large-volume organs (e.g., brainstem and mandible) but show limitations in handling complex anatomical structures (e.g., optic chiasm) and symmetrical features (e.g., bilateral optic nerves), primarily due to their constrained capacity for global contextual modeling; (2) Transformer architectures underperform consistently across all evaluation datasets, likely attributable to their inherent challenges in capturing fine-scale local features of small anatomical structures; (3) The Mamba-based framework demonstrates superior performance characteristics, particularly in maintaining the balance between local and global feature extraction.

\begin{figure}[t!]  
\centering
  \includegraphics[width=\textwidth]{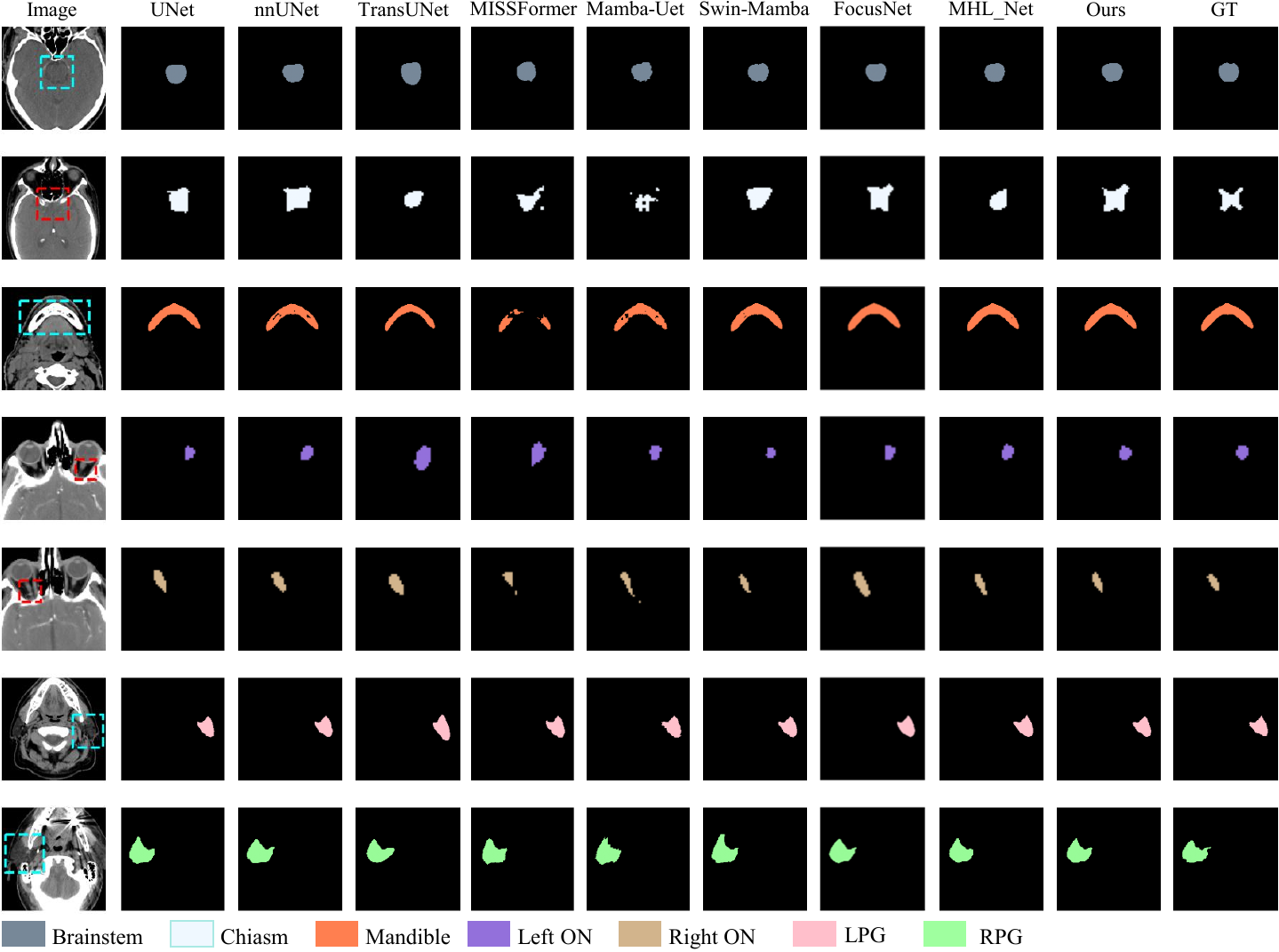} 
  \caption{Visualizations on the PDDCA dataset, where red dashed boxes indicate magnified regions and blue dashed boxes indicate non-magnified regions.}
  \label{fig:4}
\end{figure}

\begin{figure}[t!]  
\centering
  \includegraphics[width=\textwidth]{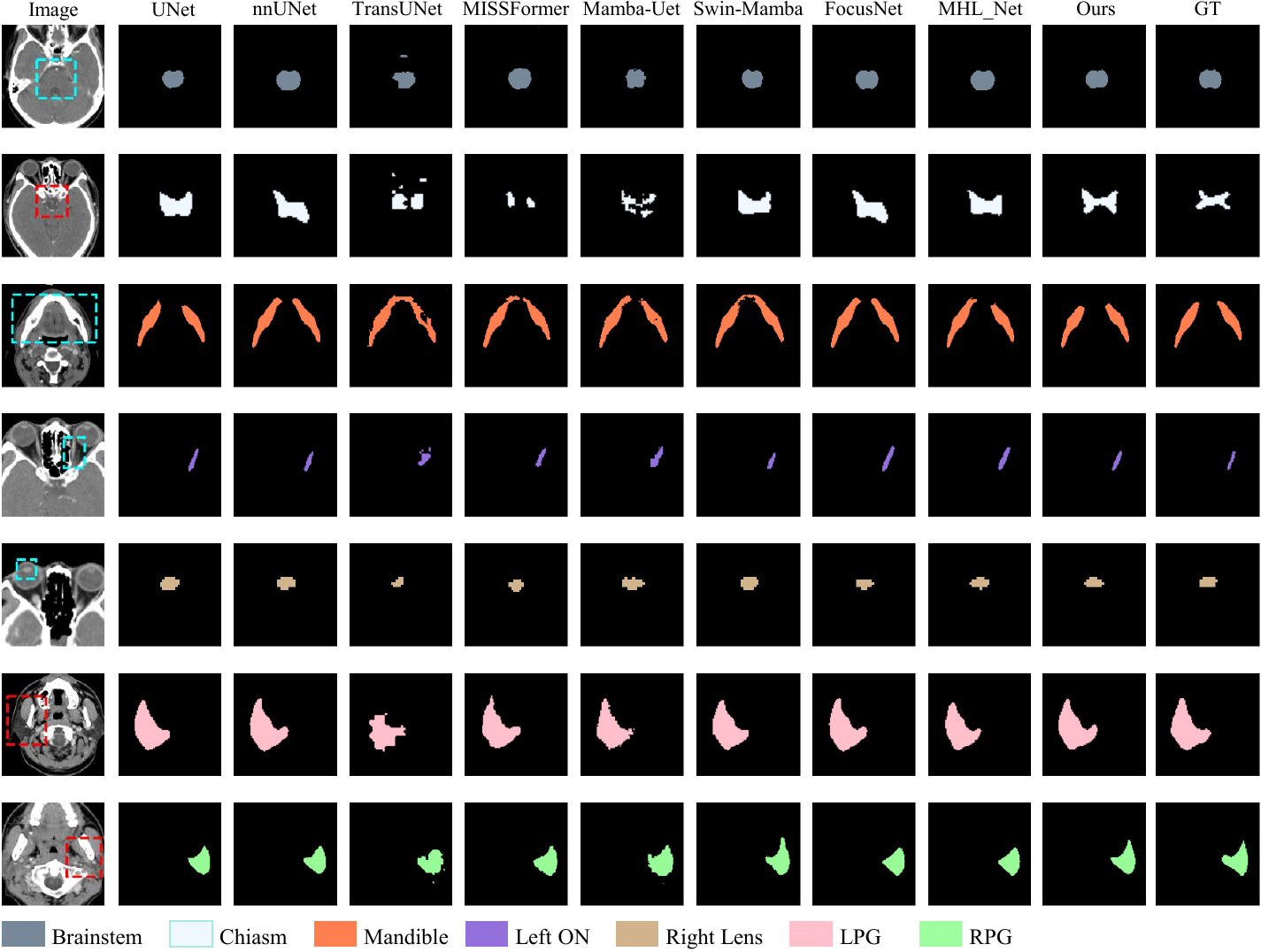} 
  \caption{Visualizations on the Structseg dataset, where red dashed boxes indicate magnified regions and blue dashed boxes indicate non-magnified regions.}
  \label{fig:5}
\end{figure}

\begin{figure}[t!]  
\centering
  \includegraphics[width=\textwidth]{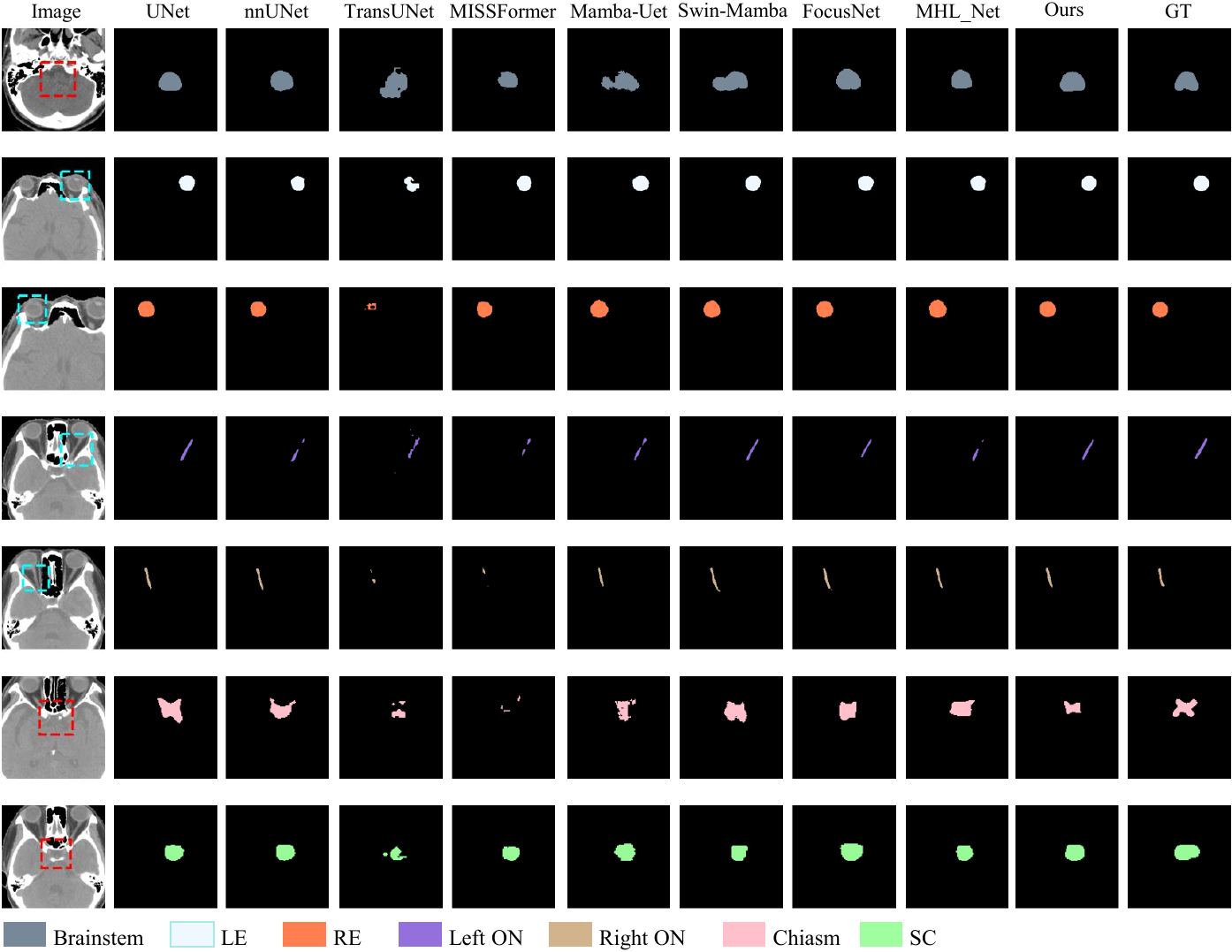} 
  \caption{Visualizations on the Inhouse dataset, where red dashed boxes indicate magnified regions and blue dashed boxes indicate non-magnified regions.}
  \label{fig:6}
\end{figure}

While two-stage approaches (FocusNet and MHL-Net) show significant improvements over single-stage methods, they still face challenges in accurately segmenting delicate structures like the chiasm (Dice score less than 60\% in PDDCA). In contrast, our method achieves state-of-the-art segmentation performance across most anatomical structures, with particular advantages in small, complex organs (average Dice improvement of 4.61\% over the second-best method in PDDCA). The qualitative results in {Figs.~\ref{fig:4}--\ref{fig:6}} further corroborate these findings, visually demonstrating our method's exceptional capability in simultaneously handling both large regular organs and small complex structures with unprecedented precision.

\begin{figure}[t!]  
\centering
  \includegraphics[width=\textwidth]{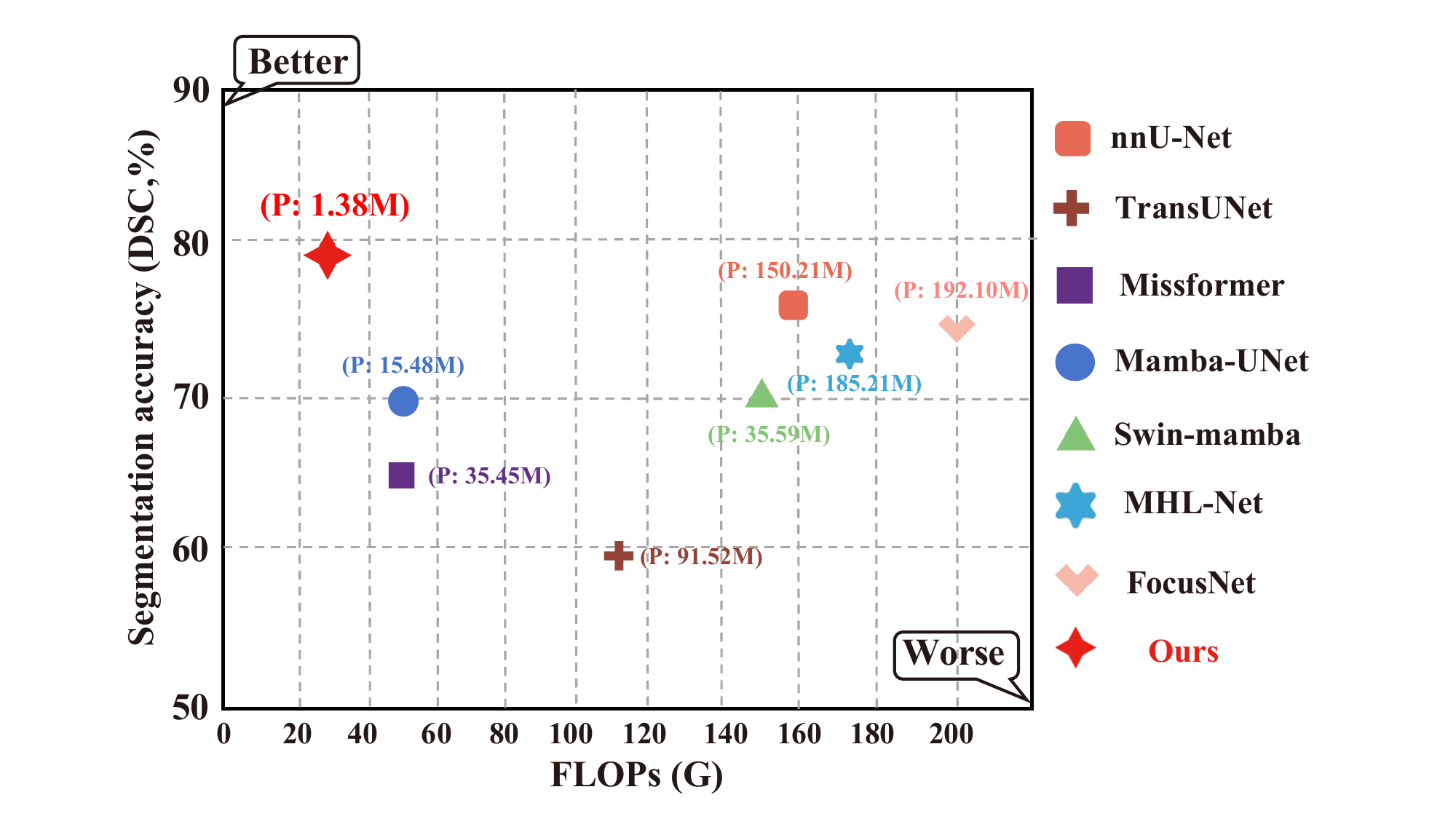} 
  \caption{Comprehensive evaluation of all compared methods.}
  \label{fig:9}
\end{figure}

{We also compared the number of parameters and computational complexity (FLOPs) of all methods under standard training configurations, which is shown in {Fig.~\ref{fig:9}}. The results show that our proposed HUR-MACL method not only achieves the best overall segmentation performance across all three datasets but also surpasses existing methods in both parameter count and computational overhead.}

\begin{table*}[t]
\centering
\caption{Segmentation results of different methods on the PDDCA dataset.}
\label{tab:1} 
\resizebox{\textwidth}{!}{%
\begin{tabular}{lcccccccc}
\hline
                                 & \multicolumn{2}{c}{\textbf{Brainstem}}                                       & \multicolumn{2}{c}{\textbf{Chiasm}}                                          & \multicolumn{2}{c}{\textbf{Mandible}}                                        & \multicolumn{2}{c}{\textbf{LON}}                                             \\ \cline{2-9} 
\multirow{-2}{*}{\textbf{PDDCA}} & \textbf{DSC$\uparrow$}                         & \textbf{ASSD$\downarrow$}                       & \textbf{DSC$\uparrow$}                         & \textbf{ASSD$\downarrow$}                       & \textbf{DSC$\uparrow$}                         & \textbf{ASSD$\downarrow$}                       & \textbf{DSC$\uparrow$}                         & \textbf{ASSD$\downarrow$}                       \\ \hline
U-Net                            & {\color[HTML]{000000} 86.99}          & {\color[HTML]{000000} 1.15}          & {\color[HTML]{000000} 46.42}          & {\color[HTML]{000000} 1.68}          & {\color[HTML]{000000} 90.23}          & {\color[HTML]{000000} 1.57}          & {\color[HTML]{000000} 71.47}          & {\color[HTML]{FE0000} \textbf{0.72}} \\
nnU-Net                          & {\color[HTML]{3531FF} \textbf{87.21}} & {\color[HTML]{3531FF} \textbf{1.03}} & {\color[HTML]{000000} 48.62}          & {\color[HTML]{000000} 1.71}          & {\color[HTML]{000000} 91.12}          & {\color[HTML]{000000} 1.42}          & {\color[HTML]{000000} 71.35}          & {\color[HTML]{000000} 0.89}          \\ \hline
TransUNet                        & 77.70 & 1.92 & 30.72 & 2.66 & 66.60 & 5.89 & 34.40 & 3.10 \\
Missformer                       & 83.20 & 1.53 & 28.87 & 3.18 & 67.31 & 5.47 & 51.95 & 3.36 \\ \hline
Mamba-UNet                       & 84.44 & 1.37 & 32.33 & 2.71 & 86.28 & 2.10 & 63.62 & 2.54 \\
Swin-mamba                       & 84.74 & 1.23 & 25.11 & 3.50 & 89.80 & 2.22 & 67.10 & 1.66 \\ \hline
MHL-Net                          & 86.75 & 1.33 & 53.16 & 1.69 & 91.22 & 1.58 & 70.03 & 0.95 \\
FocusNet                         & 86.59 & 1.55 & {\color[HTML]{3531FF} \textbf{56.42}} & {\color[HTML]{FE0000} \textbf{1.41}} & {\color[HTML]{3531FF} \textbf{92.07}} & {\color[HTML]{3531FF} \textbf{1.37}} & {\color[HTML]{3531FF} \textbf{72.46}} & 2.40 \\ \hline
\textbf{HUR-MACL}                & {\color[HTML]{FE0000} \textbf{90.07}} & {\color[HTML]{FE0000} \textbf{0.79}} & {\color[HTML]{FE0000} \textbf{66.24}} & {\color[HTML]{3531FF} \textbf{1.50}} & {\color[HTML]{FE0000} \textbf{93.99}} & {\color[HTML]{FE0000} \textbf{0.81}} & {\color[HTML]{FE0000} \textbf{76.45}} & {\color[HTML]{3531FF} \textbf{0.78}} \\ \hline
                                 & \multicolumn{2}{c}{\textbf{RON}}                                             & \multicolumn{2}{c}{\textbf{LPG}}                                             & \multicolumn{2}{c}{\textbf{RPG}}                                             & \multicolumn{2}{c}{\textbf{Average}}                                         \\ \cline{2-9} 
\multirow{-2}{*}{\textbf{PDDCA}} & \textbf{DSC$\uparrow$}                         & \textbf{ASSD$\downarrow$}                       & \textbf{DSC$\uparrow$}                         & \textbf{ASSD$\downarrow$}                       & \textbf{DSC$\uparrow$}                         & \textbf{ASSD$\downarrow$}                       & \textbf{DSC$\uparrow$}                         & \textbf{ASSD$\downarrow$}                       \\ \hline
U-Net                            & 66.88 & {\color[HTML]{FE0000} \textbf{1.11}} & 84.71 & 1.44 & 84.92 & 1.56 & 75.95 & {\color[HTML]{3531FF} \textbf{1.32}} \\
nnU-Net                          & 68.12 & 3.02 & 84.68 & 1.39 & {\color[HTML]{3531FF} \textbf{85.40}} & {\color[HTML]{3531FF} \textbf{1.48}} & 76.64 & 1.56 \\ \hline
TransUNet                        & 31.01 & 3.07 & 70.21 & 2.84 & 71.94 & 2.73 & 54.65 & 3.17 \\
Missformer                       & 47.12 & 4.07 & 76.21 & 2.48 & 76.77 & 2.42 & 61.63 & 3.22 \\ \hline
Mamba-UNet                       & 57.79 & 3.04 & 81.78 & 1.85 & 80.64 & 1.87 & 69.55 & 2.21 \\
Swin-mamba                       & 61.47 & 2.61 & 83.73 & 1.40 & 81.04 & 1.78 & 70.43 & 2.06 \\ \hline
MHL-Net                          & {\color[HTML]{3531FF} \textbf{69.15}} & 2.88 & 84.32 & 1.57 & 83.77 & 1.65 & 76.91 & 1.66 \\
FocusNet                         & 64.14 & 2.86 & {\color[HTML]{3531FF} \textbf{85.01}} & {\color[HTML]{3531FF} \textbf{1.36}} & 83.91 & 1.49 & {\color[HTML]{3531FF} \textbf{77.23}} & 1.78 \\ \hline
\textbf{HUR-MACL}                & {\color[HTML]{FE0000} \textbf{74.18}} & {\color[HTML]{3531FF} \textbf{1.80}} & {\color[HTML]{FE0000} \textbf{85.45}} & {\color[HTML]{FE0000} \textbf{1.24}} & {\color[HTML]{FE0000} \textbf{86.50}} & {\color[HTML]{FE0000} \textbf{1.17}} & {\color[HTML]{FE0000} \textbf{81.84}} & {\color[HTML]{FE0000} \textbf{1.16}} \\ \hline
\end{tabular}%
}
\end{table*}

\begin{table*}[t]
\centering
\caption{{Segmentation results of different methods on the StructSeg dataset.}}
\label{tab:2} 
\resizebox{\textwidth}{!}{%
\begin{tabular}{lcccccccc}
\hline
                                     & \multicolumn{2}{c}{\textbf{Brainstem}}                                       & \multicolumn{2}{c}{\textbf{Chiasm}}                                          & \multicolumn{2}{c}{\textbf{Mandible}}                                        & \multicolumn{2}{c}{\textbf{LON}}                                             \\ \cline{2-9} 
\multirow{-2}{*}{\textbf{StructSeg}} & \textbf{DSC$\uparrow$}                         & \textbf{ASSD$\downarrow$}                       & \textbf{DSC$\uparrow$}                         & \textbf{ASSD$\downarrow$}                       & \textbf{DSC$\uparrow$}                         & \textbf{ASSD$\downarrow$}                       & \textbf{DSC$\uparrow$}                         & \textbf{ASSD$\downarrow$}                       \\ \hline
U-Net                                & 83.04                                 & 1.57                                 & 43.60                                 & 1.87                                 & 90.38                                 & 1.54                                 & 63.56                                 & 1.00                                 \\
nnU-Net                              & 83.24                                 & 1.40                                 & 45.66                                 & 1.91                                 & {\color[HTML]{3166FF} \textbf{91.27}} & 1.40                                 & 63.45                                 & 1.24                                 \\ \hline
TransUNet                            & 77.39                                 & 1.48                                 & 39.95                                 & 3.46                                 & 75.94                                 & 2.41                                 & 45.50                                 & 1.42                                 \\
Missformer                           & 82.79                                 & 1.59                                 & 37.55                                 & 3.25                                 & 84.78                                 & 2.31                                 & 53.38                                 & 1.42                                 \\ \hline
Mamba-UNet                           & 82.96                                 & 1.71                                 & 35.39                                 & 2.14                                 & 87.79                                 & 1.94                                 & 61.35                                 & 1.14                                 \\
Swin-mamba                           & 80.83                                 & 1.63                                 & 36.26                                 & 2.41                                 & 89.83                                 & 1.56                                 & 61.66                                 & 1.07                                 \\ \hline
MHL-Net                              & {\color[HTML]{3166FF} \textbf{84.17}} & {\color[HTML]{3166FF} \textbf{1.05}} & 43.70                                 & {\color[HTML]{3166FF} \textbf{1.63}} & 88.80                                 & 1.28                                 & {\color[HTML]{3166FF} \textbf{63.72}} & {\color[HTML]{3166FF} \textbf{0.92}} \\
FocusNet                             & 84.01                                 & 1.36                                 & {\color[HTML]{3166FF} \textbf{49.65}} & 1.67                                 & 89.77                                 & {\color[HTML]{3166FF} \textbf{1.38}} & 63.71                                 & 1.34                                 \\ \hline
\textbf{HUR-MACL}                    & {\color[HTML]{FE0000} \textbf{86.54}} & {\color[HTML]{FE0000} \textbf{1.02}} & {\color[HTML]{FE0000} \textbf{53.06}} & {\color[HTML]{FE0000} \textbf{1.39}} & {\color[HTML]{FE0000} \textbf{91.41}} & {\color[HTML]{3166FF} \textbf{1.25}} & {\color[HTML]{FE0000} \textbf{72.39}} & {\color[HTML]{FE0000} \textbf{0.79}} \\ \hline
                                     & \multicolumn{2}{c}{\textbf{RON}}                                             & \multicolumn{2}{c}{\textbf{LPG}}                                             & \multicolumn{2}{c}{\textbf{RL}}                                             & \multicolumn{2}{c}{\textbf{Average}}                                         \\ \cline{2-9} 
\multirow{-2}{*}{\textbf{StructSeg}} & \textbf{DSC$\uparrow$}                         & \textbf{ASSD$\downarrow$}                       & \textbf{DSC$\uparrow$}                         & \textbf{ASSD$\downarrow$}                       & \textbf{DSC$\uparrow$}                         & \textbf{ASSD$\downarrow$}                       & \textbf{DSC$\uparrow$}                         & \textbf{ASSD$\downarrow$}                       \\ \hline
U-Net                                & 68.23                                 & 0.55                                 & 81.64                                 & 1.51                                 & 80.01                                 & 1.63                                 & 72.92                                 & 1.38                                 \\
nnU-Net                              & 69.50                                 & 1.51                                 & 81.61                                 & 1.46                                 & 80.46                                 & 1.55                                 & 73.60                                 & 1.50                                 \\ \hline
TransUNet                            & 52.34                                 & 1.12                                 & 74.79                                 & 1.92                                 & 72.25                                 & 2.10                                 & 62.59                                 & 1.99                                 \\
Missformer                           & 49.66                                 & 1.12                                 & 78.77                                 & 1.92                                 & 73.99                                 & 2.10                                 & 65.85                                 & 1.96                                 \\ \hline
Mamba-UNet                           & 67.45                                 & 0.79                                 & 78.70                                 & 1.71                                 & 77.97                                 & 1.79                                 & 70.23                                 & 1.60                                 \\
Swin-mamba                           & {\color[HTML]{3166FF} \textbf{74.46}} & {\color[HTML]{FE0000} \textbf{0.37}} & 81.34                                 & 2.14                                 & 78.84                                 & 1.72                                 & 71.89                                 & 1.56                                 \\ \hline
MHL-Net                              & 73.46                                 & {\color[HTML]{3166FF} \textbf{0.42}} & 81.38                                 & {\color[HTML]{FE0000} \textbf{1.34}} & 79.84                                 & {\color[HTML]{3166FF} \textbf{1.44}} & 73.58                                 & {\color[HTML]{3166FF} \textbf{1.15}} \\
FocusNet                             & 68.01                                 & 1.05                                 & {\color[HTML]{3166FF} \textbf{84.27}} & 1.46                                 & {\color[HTML]{3166FF} \textbf{82.30}} & 1.69                                 & {\color[HTML]{3166FF} \textbf{74.53}} & 1.42                                 \\ \hline
\textbf{HUR-MACL}                    & {\color[HTML]{FE0000} \textbf{75.62}} & {\color[HTML]{3166FF} \textbf{0.43}} & {\color[HTML]{FE0000} \textbf{85.29}} & {\color[HTML]{3166FF} \textbf{1.28}} & {\color[HTML]{FE0000} \textbf{84.00}} & {\color[HTML]{FE0000} \textbf{1.40}} & {\color[HTML]{FE0000} \textbf{78.32}} & {\color[HTML]{FE0000} \textbf{1.08}} \\ \hline
\end{tabular}%
}
\end{table*}

\begin{table*}[t]
\centering
\caption{Segmentation results of different methods on the Inhouse dataset.}
\label{tab:3} 
\resizebox{\textwidth}{!}{%
\begin{tabular}{lcccccccc}
\hline
                                   & \multicolumn{2}{c}{\textbf{Brainstem}}                                       & \multicolumn{2}{c}{\textbf{Chiasm}}                                          & \multicolumn{2}{c}{\textbf{SC}}                                              & \multicolumn{2}{c}{\textbf{LON}}                                             \\ \cline{2-9} 
\multirow{-2}{*}{\textbf{Inhouse}} & \textbf{DSC$\uparrow$}                         & \textbf{ASSD$\downarrow$}                       & \textbf{DSC$\uparrow$}                         & \textbf{ASSD$\downarrow$}                       & \textbf{DSC$\uparrow$}                         & \textbf{ASSD$\downarrow$}                       & \textbf{DSC$\uparrow$}                         & \textbf{ASSD$\downarrow$}                       \\ \hline
U-Net                              & 81.41 & 2.67 & 40.99 & {\color[HTML]{FE0000} \textbf{2.77}} & 73.13 & 1.28 & {\color[HTML]{3531FF} \textbf{63.60}} & {\color[HTML]{3531FF} \textbf{1.68}} \\
nnU-Net                            & {\color[HTML]{3531FF} \textbf{81.61}} & {\color[HTML]{3531FF} \textbf{2.39}} & {\color[HTML]{3531FF} \textbf{41.09}} & 2.48 & {\color[HTML]{FE0000} \textbf{73.31}} & {\color[HTML]{FE0000} \textbf{1.15}} & 63.44 & {\color[HTML]{FE0000} \textbf{1.51}} \\ \hline
TransUNet                          & 75.70 & 3.43 & 32.46 & 3.26 & 62.12 & 2.06 & 43.96 & 3.62 \\
Missformer                         & 79.24 & 2.92 & 33.16 & 3.77 & 67.98 & 1.52 & 57.91 & 2.35 \\ \hline
Mamba-UNet                         & 79.63 & 2.80 & 35.91 & 3.12 & 69.85 & 1.65 & 61.00 & 1.88 \\
Swin-mamba                         & 75.24 & 3.10 & 36.06 & {\color[HTML]{3531FF} \textbf{2.95}} & 68.42 & 1.61 & 61.70 & 1.96 \\ \hline
MHL-Net                            & 79.62 & 2.43 & 37.52 & 3.40 & 69.05 & 1.25 & 58.92 & 1.99 \\
FocusNet                           & 78.80 & 2.46 & 40.21 & 3.16 & 69.94 & 2.01 & 57.39 & 2.12 \\ \hline
\textbf{HUR-MACL}                  & {\color[HTML]{FE0000} \textbf{82.59}} & {\color[HTML]{FE0000} \textbf{2.34}} & {\color[HTML]{FE0000} \textbf{44.55}} & 3.08 & {\color[HTML]{3531FF} \textbf{73.23}} & {\color[HTML]{3531FF} \textbf{1.18}} & {\color[HTML]{FE0000} \textbf{66.08}} & 1.85 \\ \hline
                                   & \multicolumn{2}{c}{\textbf{RON}}                                             & \multicolumn{2}{c}{\textbf{LE}}                                              & \multicolumn{2}{c}{\textbf{RE}}                                              & \multicolumn{2}{c}{\textbf{Average}}                                         \\ \cline{2-9} 
\multirow{-2}{*}{\textbf{Inhouse}} & \textbf{DSC$\uparrow$}                         & \textbf{ASSD$\downarrow$}                       & \textbf{DSC$\uparrow$}                         & \textbf{ASSD$\downarrow$}                       & \textbf{DSC$\uparrow$}                         & \textbf{ASSD$\downarrow$}                       & \textbf{DSC$\uparrow$}                         & \textbf{ASSD$\downarrow$}                       \\ \hline
U-Net                              & 64.15 & 2.00 & 91.13 & 1.15 & 91.99 & 1.11 & 72.34 & 1.81 \\
nnU-Net                            & 64.31 & {\color[HTML]{3531FF} \textbf{1.79}} & 90.98 & {\color[HTML]{3531FF} \textbf{1.03}} & {\color[HTML]{3531FF} \textbf{92.22}} & 1.00 & {\color[HTML]{3531FF} \textbf{72.42}} & {\color[HTML]{FE0000} \textbf{1.62}} \\ \hline
TransUNet                          & 48.82 & 2.57 & 88.11 & 1.58 & 86.93 & 1.77 & 62.59 & 2.61 \\
Missformer                         & 55.58 & 2.26 & 90.39 & 1.31 & 89.73 & 1.40 & 67.71 & 2.22 \\ \hline
Mamba-UNet                         & 62.68 & 1.80 & 91.20 & 1.21 & 90.61 & 1.52 & 70.13 & 2.00 \\
Swin-mamba                         & 60.30 & 1.85 & {\color[HTML]{3531FF} \textbf{91.33}} & 1.15 & 91.52 & 1.09 & 69.22 & 1.96 \\ \hline
MHL-Net                            & 62.36 & 1.80 & 89.67 & 1.08 & 89.97 & {\color[HTML]{3531FF} \textbf{0.99}} & 69.59 & 1.85 \\
FocusNet                           & {\color[HTML]{3531FF} \textbf{66.09}} & 1.82 & 89.97 & 1.13 & 92.00 & 1.09 & 70.63 & 1.97 \\ \hline
\textbf{HUR-MACL}                  & {\color[HTML]{FE0000} \textbf{67.80}} & {\color[HTML]{FE0000} \textbf{1.65}} & {\color[HTML]{FE0000} \textbf{92.09}} & {\color[HTML]{FE0000} \textbf{1.01}} & {\color[HTML]{FE0000} \textbf{92.40}} & {\color[HTML]{FE0000} \textbf{0.95}} & {\color[HTML]{FE0000} \textbf{74.11}} & {\color[HTML]{3531FF} \textbf{1.72}} \\ \hline
\end{tabular}%
}
\end{table*}

\begin{table}[t]
\centering
\caption{{Ablation results (DSC) of different modules proposed in our method.}}
\label{tab:4} 
\resizebox{0.7\textwidth}{!}{%
\begin{tabular}{ccccccc}
\hline
ViM & Transformer & DCNN & HFD & PDDCA & StructSeg & Inhouse \\ \hline
    &      &     &     & 77.69 & 74.36     & 69.02   \\
\checkmark &      &     &     & 80.83 & 75.78     & 71.36   \\
    & \checkmark&     &     & 79.44		 & 75.03     &70.13 \\
  &  &     \checkmark&    &  80.30 & 75.18     & 70.07   \\
\checkmark & &  \checkmark&     & 81.66 & 77.21     & 73.71   \\
\checkmark &    &  \checkmark & \checkmark & \color[HTML]{FE0000} \textbf{81.84} & \color[HTML]{FE0000} \textbf{78.32}     & \color[HTML]{FE0000} \textbf{74.11}   \\ \hline
\end{tabular}
}
\end{table}

\begin{figure}[t!]  
\centering
  \includegraphics[width=\textwidth]{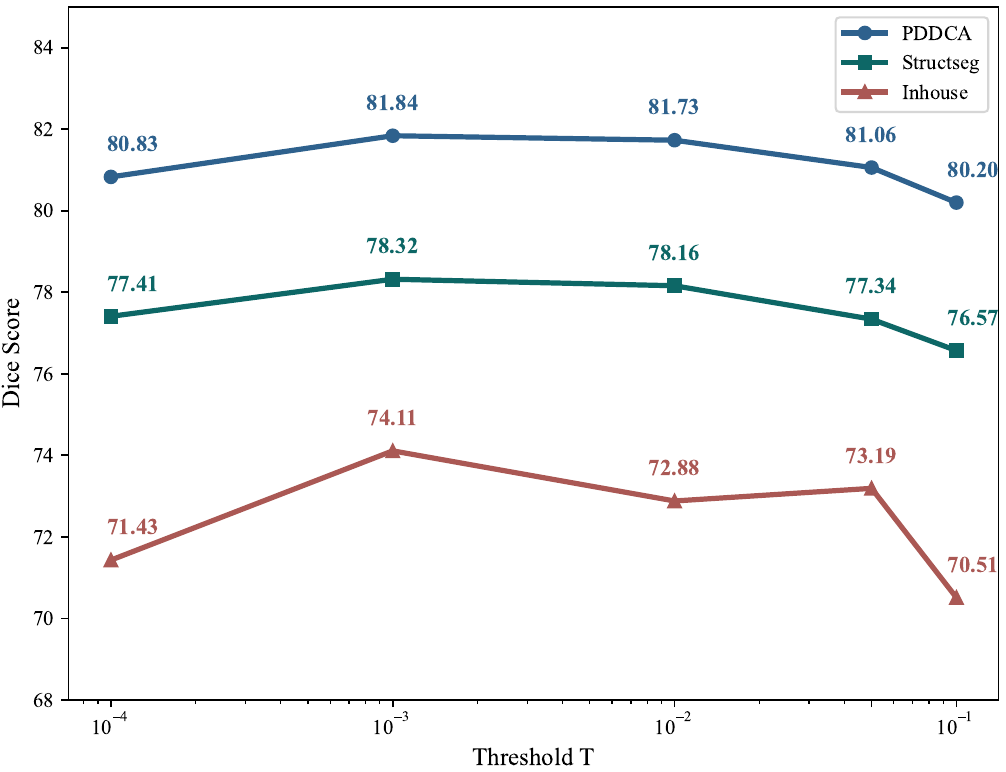} 
  \caption{DSC values for different threshold $T$.}
  \label{fig:7}
\end{figure}

\begin{figure}[t!]  
\centering
  \includegraphics[width=\textwidth]{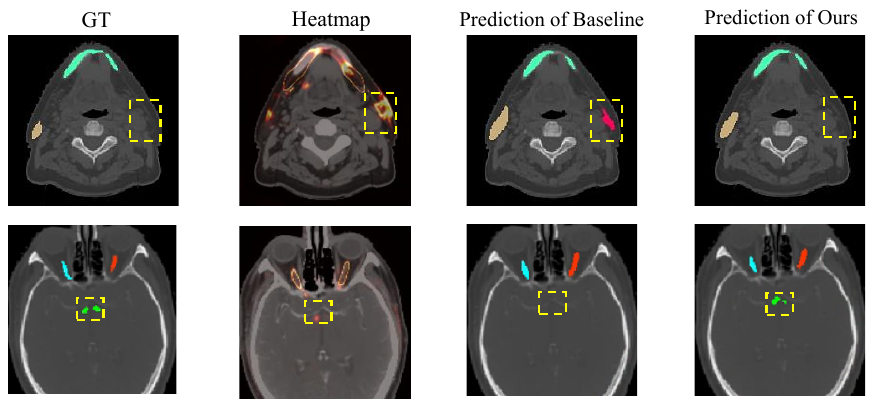} 
  \caption{Visualizations from the uncertainty heatmap, prediction of Baseline method, and prediction of our method.}
  \label{fig:8}
\end{figure}

\subsection{Ablation Study}
\label{sec44}
We performed a systematic ablation study to assess the individual contributions of each proposed component, with quantitative results presented in {Table~\ref{tab:4}}. {The evaluation was performed by progressively enhancing our baseline model—a 4-level U-Net with all proposed modules removed  with a fixed random seed (seed=42), as indicated in the first row—through the sequential integration of: (1) the dual-branch segmentation architecture that combines Vision Mamba (ViM), Transformer, and Deformable Convolutional Neural Network (DCNN), and (2) the Heterogeneous Feature Distillation (HFD) loss function.}

The incorporation of our multi-architecture segmentation framework yielded substantial performance improvements, with the ViM branch demonstrating particular advantages in processing anatomically complex regions (e.g., achieving a 1.29\% higher Dice score than DCNN in Inhouse). {This performance gap highlights ViM's superior capability in modeling long-range dependencies.}

The addition of HFD loss further enhanced model performance on average across all evaluation metrics in three datasets. This improvement stems from the loss function's effective facilitation of bidirectional feature-level knowledge transfer between the ViM and DCNN branches. The distillation process enables each architecture to compensate for the other's limitations-while DCNN provides robust local feature extraction, ViM contributes global contextual understanding, particularly benefiting challenging segmentation targets such as thin neural structures and low-contrast tissue boundaries.

\subsection{Sensitivity Analysis of Threshold $T$}
We further investigated the impact of the uncertainty-guided threshold parameter $T$ on the final segmentation performance. The model was evaluated using five different values of $T$: $\{0.1, 0.05, 0.01, 0.001, 0.0001\}$, across all three datasets.  
As depicted in {Fig.~\ref{fig:7}}, the optimal performance was achieved at $T = 0.001$, yielding the highest average DSC across datasets. A threshold smaller than this, specifically $T = 0.0001$, resulted in performance degradation, with a notable 2.68\% decrease decrease in DSC on the Inhouse dataset. This suggests that excessively lowering the threshold causes overexpansion of the inclusion region, leading to unnecessary feature fusion and functional overlap between model components, which ultimately impairs segmentation accuracy.  
These results highlight the importance of a balanced $T$ value to effectively guide the interaction between architecture-specific pathways without causing excessive redundancy.

\subsection{Visualization of Features}
We conducted a qualitative evaluation by visualizing feature outputs from the baseline model, uncertainty estimation maps, and our proposed method, with representative examples presented in {Fig.~\ref{fig:8}}. The uncertainty maps consistently highlight regions of high prediction uncertainty, particularly in anatomically complex areas (e.g., parotid gland boundaries), low-contrast structures (e.g., optic chiasm), and symmetric organs, confirming our hypothesis that these clinically challenging regions are most susceptible to segmentation errors. Compared to the baseline, which exhibits characteristic failures such as left parotid gland misclassification and chiasm under-segmentation, our method demonstrates substantially improved accuracy through uncertainty-guided adaptive learning. By dynamically modulating feature fusion based on localized uncertainty, our approach achieves more precise boundary delineation and significantly reduces mis-segmentation artifacts. These visual observations strongly corroborate our quantitative results, providing compelling evidence for the efficacy of our hybrid architecture in overcoming key challenges in medical image segmentation.

{\subsection{Cross-Dataset Experimental Results}}

{Table ~\ref{tab:dsc_performance} below presents the DSC (\%) performance of all methods on the Inhouse test set under the cross-dataset setting (trained on public dataset $\rightarrow$ tested on private Inhouse dataset). As shown in the table, under the more challenging cross-dataset scenario (trained on a single public dataset only), our proposed HUR-MACL model achieves the best performance across all four organs and in the average DSC. This validates that our method maintains superior segmentation capability even when there is a significant distribution shift between the training and testing data, demonstrating its good generalizability and robustness.}
\begin{table}[htbp]
  \centering
  \caption{{DSC (\%) performance of all methods on the Inhouse test set under the cross-dataset setting.}}
  \label{tab:dsc_performance}
  \begin{tabular}{lccccc}
    \toprule
    Method & Brainstem & Chiasm & LON & RON & Average \\
    \midrule
    U-Net       & 78.9 & 36.5 & 60.2 & 60.8 & 59.1 \\
    nnUNet      & 79.3 & 37.8 & 61.5 & 62.1 & 60.2 \\
    \midrule
    TransUNet   & 73.2 & 29.4 & 41.5 & 46.7 & 47.7 \\
    Missformer  & 76.5 & 30.8 & 54.9 & 53.8 & 54.0 \\
    \midrule
    Mamba-UNet  & 77.8 & 33.5 & 59.0 & 60.5 & 57.7 \\
    Swin-Mamba  & 73.9 & 33.8 & 59.5 & 58.4 & 56.4 \\
    \midrule
    MHL-Net     & 77.5 & 34.2 & 56.8 & 60.2 & 57.2 \\
    FocusNet    & 77.1 & 37.0 & 55.3 & 64.1 & 58.4 \\
    \midrule
    \textbf{HUR-MACL (Ours)} & \textbf{80.5} & \textbf{40.8} & \textbf{63.5} & \textbf{65.4} & \textbf{62.6} \\
    \bottomrule
  \end{tabular}
\end{table}

\section{Discussions}
In this paper, we proposed a new hybrid architecture model, which addresses the critical challenge of multi-organ segmentation in the head and neck region by integrating high uncertainty region mining, multi-architecture collaboration, and heterogeneous feature distillation. This approach adaptively identifies challenging regions and leverages the complementary strengths of Vision Mamba and Deformable CNN to enhance segmentation accuracy. The model’s design effectively mitigates the limitations of existing methods, while achieving state-of-the-art performance across multiple datasets.

The experimental results demonstrate that HUR-MACL outperforms eight comparative methods on two public datasets (PDDCA and StructSeg) and one private dataset (Inhouse), particularly for small and complex organs like the optic chiasm and spinal cord. For instance, on the PDDCA dataset, HUR-MACL achieves a DSC of 66.24\% for the chiasm, significantly higher than other methods. The visualizations further confirm the model’s ability to refine contours and reduce misclassification in challenging regions, underscoring its clinical applicability.

The results demonstrate that CNN-based models perform well on large, regularly shaped organs but poorly on small organs, indicating that conventional convolutional networks are sufficient for segmenting large regular structures without requiring complex architectures. In contrast, transformer-based methods underperform across all three datasets, while Mamba-based approaches surpass transformers, suggesting Mamba's superior feature extraction capability. Although two-stage methods achieve better performance on small structures, their overall effectiveness still lags behind our proposed method, likely due to the challenges of simultaneously optimizing multiple networks to achieve optimal performance.

The success of HUR-MACL can be attributed to three key innovations. First, the high uncertainty region mining module dynamically identifies areas requiring additional attention. Second, the collaborative use of ViM and DCNN combines global feature capture with adaptive local feature extraction, addressing the limitations of single-architecture models. Third, the heterogeneous feature distillation loss facilitates knowledge exchange between architectures, enhancing performance in hard regions. These components work synergistically to improve segmentation accuracy.

Despite its advantage, HUR-MACL has certain limitations. The model’s performance heavily relies on the threshold parameter T for hard region identification, which requires careful tuning. Additionally, the integration of multiple architectures increases model complexity, potentially limiting its deployment in resource-constrained environments. Furthermore, the current evaluation is limited to head and neck datasets, and the model’s generalizability to other anatomical regions remains to be validated. Future work could explore adaptive thresholding mechanisms and lightweight variants to address these challenges.

In the future, HUR-MACL’s framework holds promise for broader applications in medical image segmentation. The principles of hard region guidance and multi-architecture collaboration could be adapted to other complex segmentation tasks, such as abdominal or thoracic organ segmentation. Moreover, the model’s emphasis on feature distillation and adaptive learning could inspire advancements in collaborative learning for multi-modal or multi-center datasets. By continuing to refine its components and expand its scope, HUR-MACL could become a versatile tool for advancing precision in medical image analysis.

\section{Conclusion}
In this paper, we propose a novel high uncertainty guided multi-architecture collaborative learning framework (HUR-MACL) for head and neck multi-organ segmentation. Incorporating high uncertainty region mining, multi-architecture collaborative segmentation, and heterogeneous feature distillation strategies, the model adaptively identifies hard regions and effectively integrates the advantage of multiple architectures. Extensive results demonstrate that our method significantly outperforms comparative methods, providing a new perspective in this field.

\section*{Acknowledgement}
This work was supported by National Natural Science Foundation of China under Grant 82072021.

\bibliography{ref}

\end{document}